\documentclass[10pt,twocolumn,letterpaper]{article}

\usepackage{iccv}
\usepackage{times}
\usepackage{epsfig}
\usepackage{graphicx}
\usepackage{amsmath}
\usepackage{amssymb}

\usepackage{array}
\usepackage{booktabs}
\usepackage{multirow}
\usepackage{colortbl}
\usepackage{enumitem}

\usepackage[breaklinks=true,bookmarks=false]{hyperref}

\iccvfinalcopy 

\def\iccvPaperID{****} 

\ificcvfinal\fi

\begin{document}

\title{Cyclic-Bootstrap Labeling for Weakly Supervised Object Detection}

\author{Yufei~Yin$^{1}$~~~~~Jiajun~Deng$^{2}$~~~~~Wengang~Zhou$^{1,3,}$\thanks{Corresponding authors: Wengang Zhou and Houqiang Li}~~~~~Li~Li$^{1}$~~~~~Houqiang~Li$^{1,3,}$\footnotemark[1] \\
	{\normalsize $^{1}$ CAS Key Laboratory of Technology in GIPAS, EEIS Department, University of Science and Technology of China} \\
        {\normalsize $^{2}$ The University of Sydney} \\
	{\normalsize $^{3}$ Institute of Artificial Intelligence, Hefei Comprehensive National Science Center} \\
	{\tt\small yinyufei@mail.ustc.edu.cn, jiajun.deng@sydney.edu.au, \{zhwg,lil1,lihq\}@ustc.edu.cn}
}

\maketitle
\ificcvfinal\thispagestyle{empty}\fi

\begin{abstract}
Recent progress in weakly supervised object detection is featured by a combination of multiple instance detection networks (MIDN) and ordinal online refinement. However, with only image-level annotation, MIDN inevitably assigns high scores to some unexpected region proposals when generating pseudo labels. These inaccurate high-scoring region proposals will mislead the training of subsequent refinement modules and thus hamper the detection performance. In this work, we explore how to ameliorate the quality of pseudo-labeling in MIDN. Formally, we devise Cyclic-Bootstrap Labeling (CBL), a novel weakly supervised object detection pipeline, which optimizes MIDN with rank information from a reliable teacher network. Specifically, we obtain this teacher network by introducing a weighted exponential moving average strategy to take advantage of various refinement modules.  A novel class-specific ranking distillation algorithm is proposed to leverage the output of weighted ensembled teacher network for distilling MIDN with rank information.  As a result, MIDN is guided to assign higher scores to accurate proposals among their neighboring ones, thus benefiting the subsequent pseudo labeling. Extensive experiments on the prevalent PASCAL VOC 2007 \& 2012 and COCO datasets demonstrate the superior performance of our CBL framework. Code will be available at \url{https://github.com/Yinyf0804/WSOD-CBL/}.
\end{abstract}

\begin{figure}[t!]
	\centering
    \includegraphics[scale = 0.28]{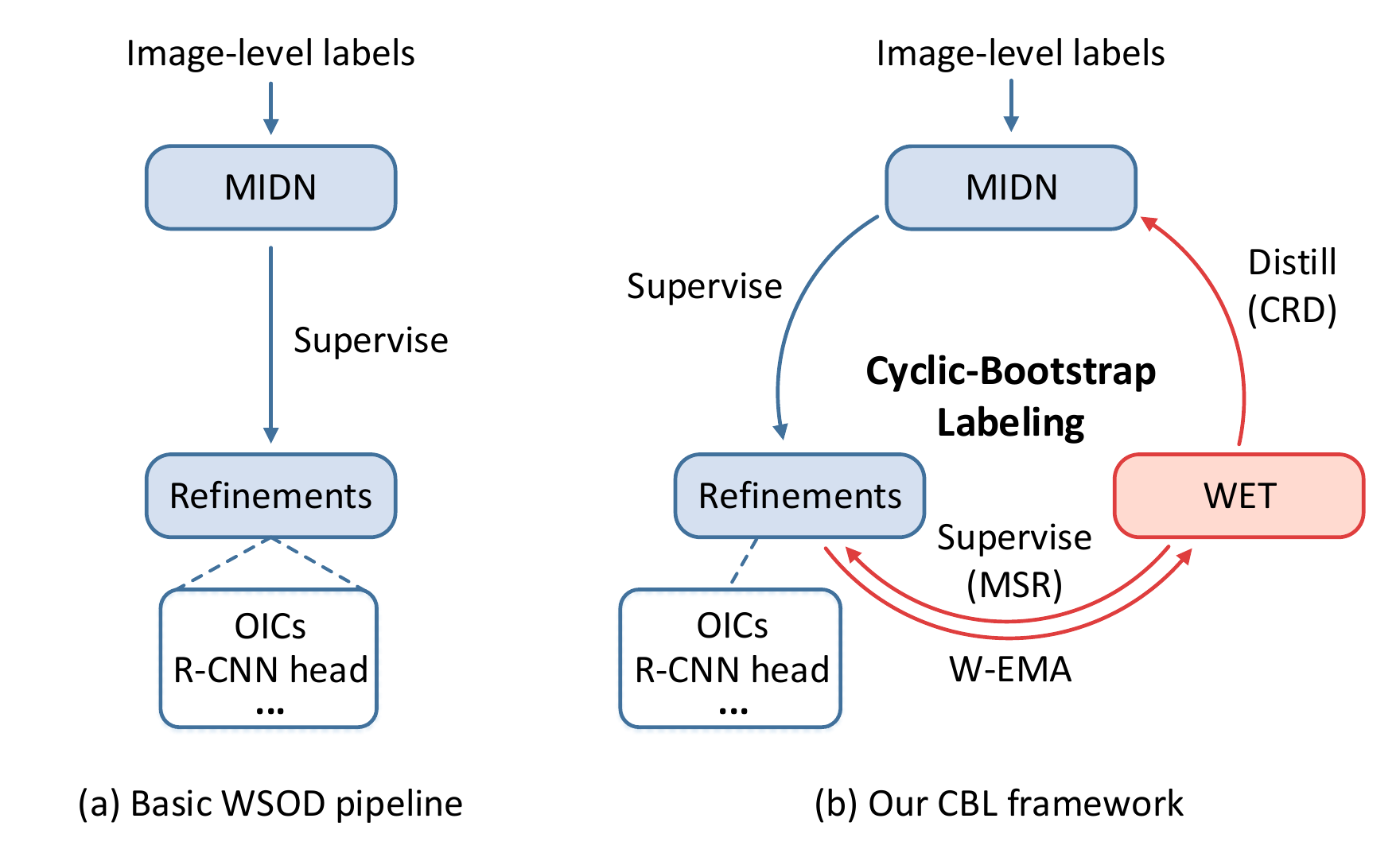} \\
	\caption{Comparison between the basic WSOD pipeline and our proposed CBL framework (feature extractor is omitted for simplicity). In the CBL framework, the subsequent refinement modules of MIDN are finally utilized to distill MIDN in turn, forming a cyclic-bootstrap procedure.} 
        \vspace{-0.2in}
	\label{wsod}
\end{figure}

\section{Introduction}
With the rapid advancements in deep neural networks, object detection has experienced significant progress. Nevertheless, state-of-the-art object detection methods are contingent upon accurate instance-level annotations obtained through fully-supervised learning settings. The process of collecting such annotations is both arduous and costly. Such facts motivate the exploration of weakly supervised object detection (WSOD) \cite{WSDDN, OICR, Yang_2019_ICCV, C-midn, WSOD2, MIST}, which achieves the object detection task using only image-level labels.

Most existing WSOD approaches in the literature  \cite{OICR, Yang_2019_ICCV} generally follow the training pipeline of Fig.~\ref{wsod}(a). First, a multiple instance detection network (MIDN)~\cite{WSDDN} is obtained by leveraging multiple instance learning scheme for optimization, which converts WSOD into a multi-class classification problem over bottom-up generated region proposals \cite{SS}. The region proposals with high scores out of MIDN are then exploited to generate pseudo ground-truth boxes, which are used for the training of refinement modules, \textit{e.g.}, online instance classifiers (OICs), and regressors (R-CNN head). By introducing MIL to the training process, MIDN obtains the capability of estimating whether an object exists in the corresponding region. However, many high-scoring region proposals from MIDN only cover the discriminative part of an instance (\textit{e.g.}, the head of a bird), or contain some background regions~\cite{OICR, WSOD2}. The improper scoring assignment of MIDN leads to the generation of inaccurate pseudo ground-truth boxes, which further hinder the training of subsequent refinement modules.

This problem has been recognized by the community, and several approaches have been proposed to address it. In the literature, C-MIDN~\cite{C-midn} and P-MIDN~\cite{PMIDN} design complementary MIDN modules to find the remained discriminative parts other than the top-scoring ones.  IM-CFB \cite{IMCFB} develops a class feature bank to collect intra-class diversity information, and devises an FGIM algorithm to ameliorate the region proposal selection.  These approaches primarily concentrate on the issue that the high-scoring proposals cover  only the most discriminative parts, while struggling to address other inaccurate scoring-assignment issues.  Besides, most of these attempts involve an auxiliary model, without offering assistance in the training of MIDN.

To this end, in this paper, we propose the Cyclic-Bootstrap Labeling (CBL) framework, which advances WSOD from the one-way pipeline (see Fig.~\ref{wsod} (a)) to a cyclic-bootstrap procedure (see Fig.~\ref{wsod} (b)). As shown in Figure, the subsequent modules of MIDN are eventually utilized to enhance itself.  In comparison to the prior approaches, CBL exerts an additional rank-based supervision on MIDN, which is capable of handling a broader  set of inaccurate scoring-assignment cases.

Specifically, following the common practice of WSOD, we first employ MIDN to generate pseudo labels, which serve as the initial supervision for subsequent refinement modules. To obtain more accurate classification results, we construct a weighted ensemble teacher (WET) model, inspired by the success of mean teacher methods \cite{tarvainen2017mean, liu2021unbiased}. The WET model is updated through the weighted exponential moving average (W-EMA) strategy, which takes advantage of multiple student candidates in the refinement modules. Subsequently, leveraging the WET results, we propose a class-specific ranking distillation (CRD) algorithm to supervise MIDN with rank-based labels in a distillation manner. This additional rank-based supervision allows MIDN to achieve an improved scoring (ranking) assignment, where accurate proposals will be assigned higher scores among their neighboring ones. Moreover, we observe that the WET model can also act as a reliable teacher for the  R-CNN head in the basic WSOD pipeline \cite{Yang_2019_ICCV}. To this end, we propose a multi-seed R-CNN (MSR) algorithm to mine multiple positive seeds according to the WET results, calculate their confidence scores, and utilize them to generate pseudo labels for the supervision of the  R-CNN head.

Our main contributions are summarized as follows:
\begin{itemize} [noitemsep,nolistsep]
    \item We propose a novel cyclic-bootstrap labeling (CBL) framework for weakly supervised object detection. The proposed  CBL contains a weighted ensemble teacher model to generate reliable detection results, a class-specific ranking distillation algorithm to distill the MIDN module with rank information, and a multi-seed R-CNN algorithm to mine accurate positive seeds for the training of the  R-CNN head.
    \item We provide a new perspective that the subsequent modules of MIDN are finally utilized to distill MIDN in turn, forming a cyclic-bootstrap procedure, which is rarely explored in previous WSOD works.
    \item Extensive experiments on the prevalent PASCAL VOC 2007 \& 2012 and COCO datasets demonstrate the superior performance of our CBL framework.

\end{itemize}

\section{Related Work}
In this section, we briefly review the related methods including Weakly supervised object detection (WSOD) and Knowledge distillation.
\subsection{Weakly Supervised Object Detection}
Weakly supervised object detection (WSOD) \cite{WSDDN, ContextLocNet, OICR, PCL, C-WSL, WS-JDS, Zigzag, ts2c, WSCCN, SDCN, Selftaught, WSRPN, MELM, PredNet, C-midn, c-mil, Yang_2019_ICCV, WSOD2, gao2019utilizing, OIM, MIST, SLV, objectness, IMCFB, PMIDN, zhang2021weakly, cheng2020high, zhang2020weakly, feng2020tcanet, shen2020enabling, yin2022fi, CASD, seo2022object, liao2022end, huang2022w2n, SoS} has attracted much attention in recent years. Most recent works utilize Multiple Instance Learning (MIL) strategy to convert WSOD into a multi-class classification task, and adopt WSDDN \cite{WSDDN} as the basic multiple instance detection network (MIDN) in their frameworks.  WSDDN adopts MIL into a CNN network with a two-stream structure  (\textit{i.e.}, classification stream and detection stream), and combines the scores obtained from these two streams to generate instance-level scores. To improve the detection capability,  on one hand, some works add several modules based on WSDDN for online refinement. OICR \cite{OICR} first adds several cascaded online instance classifiers to refine the classification results, and adopts a top-scoring strategy to obtain pseudo seeds for training these classifiers. To obtain more accurate seeds, WSOD$^2$ \cite{WSOD2} adopts bottom-up object evidence to update the original classification score during selection, \cite{ts2c} and \cite{objectness} utilize results from the other tasks for assistance and MIST \cite{MIST} proposes a multiple instance self-training algorithm. Furthermore, Yang \cite{Yang_2019_ICCV} constructs a multi-task rcnn-head to adjust the positions and shapes of proposals.

On the other hand, some works propose to improve the basic MIDN. C-MIDN and P-MIDN \cite{C-midn, PMIDN} design a (several) complementary MIDN module(s) to find the remained object parts other than the top-scoring one, WS-JDS \cite{WS-JDS} introduces the segmentation task for assistance, and IM-CFB \cite{IMCFB} constructs a class feature bank to collect intra-class diversity information for amelioration. This work also aims to improve MIDN, but different from them, we propose to re-adjust the rank distribution of MIDN among neighboring positive instances, thus helping to generate more high-quality pseudo labels for subsequent refinement.

\subsection{Knowledge Distillation}
Knowledge distillation is firstly proposed in \cite{hinton2015distilling} to transfer knowledge from complicated teacher models to distilled student models, which makes the students achieve similar performance to that of teachers. Knowledge distillation has been explored by a series of works in different tasks \cite{chen2017learning, wang2019distilling, guo2021distilling, li2022knowledge, xu2022vitpose, wang2023biased}. \cite{chen2017learning} propose two new losses for better knowledge distillation on the classification and regression task, and combine hint learning to help the training process. \cite{wang2019distilling} employ the inter-location discrepancy of teacher's feature response on near object anchor locations for knowledge distillation. \cite{guo2021distilling} use KL divergence loss to distill the classification head while processing proposals and negative proposals separately. \cite{li2022knowledge} apply a similar  rank distillation strategy with our CRD algorithm, while their main difference is that our work focuses on the WSOD task which does not have instance-level labels. To address this problem, we generate a reliable positive proposal set based on the more accurate WET results and propose a weighted KL divergence loss to alleviate the negative effects brought by noisy labels.

Previous WSOD work SoS \cite{SoS} also adopts the knowledge distillation strategy. 
However, the usage in \cite{SoS} simply follows the semi-supervised object detection paradigm and  will not ameliorate the original WSOD network. In contrast, our method adopts data distillation in the WSOD training procedure to improve the rank distribution of MIDN, thus benefitting the whole WSOD network. 

\begin{figure*}[t!]
	\centering
    \includegraphics[width=1.0\textwidth,height=0.61\textwidth]{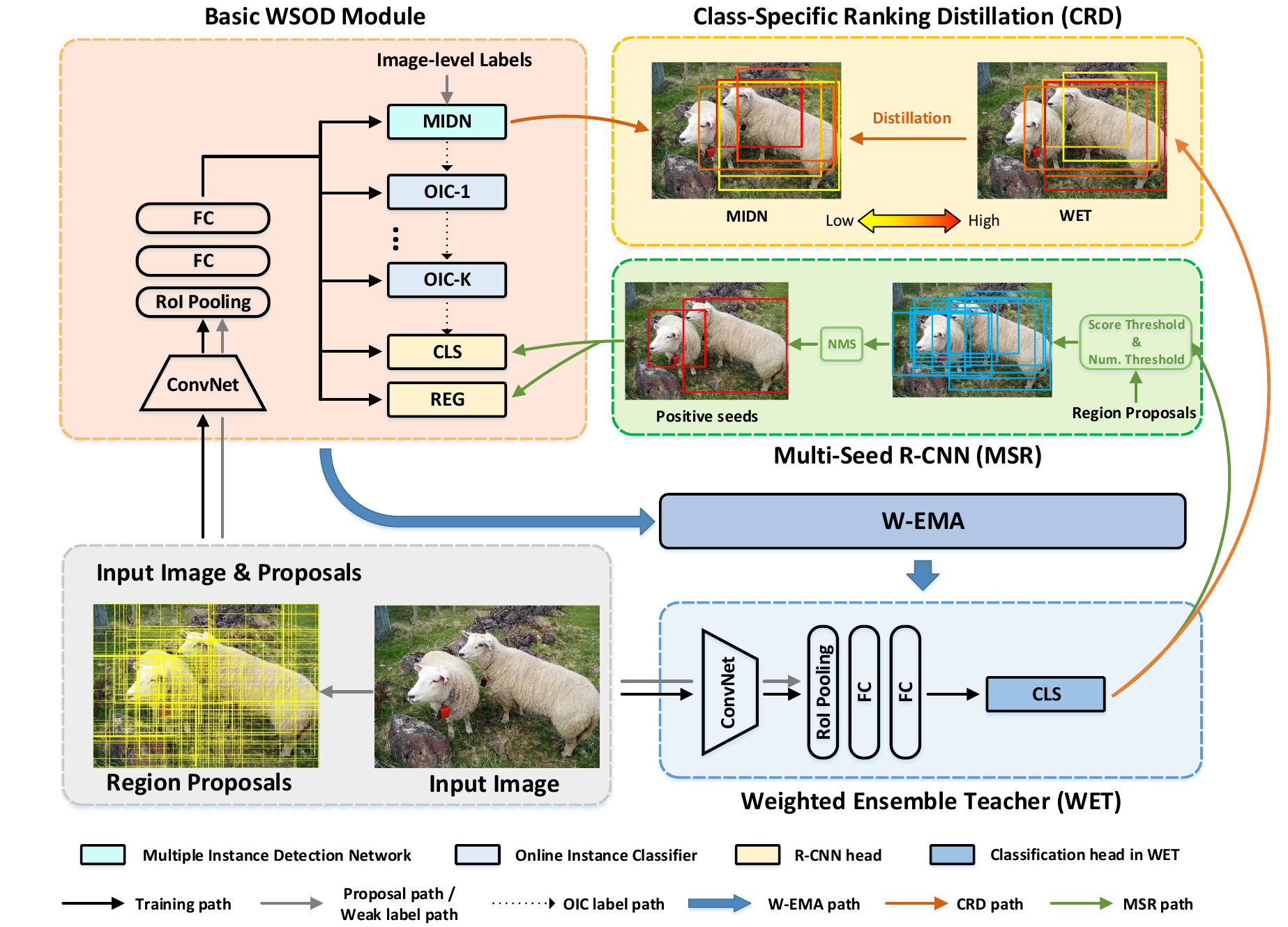}
	\vspace{-8pt}
    \vspace{-0.05in}
	\caption{An overview of our CBL framework. Proposal features are first fed into the MIDN module to produce instance-level scores. 
    Meanwhile, the image and corresponding proposals are sent to the weighted ensemble teacher (WET) model, which is gradually updated by the basic WSOD module via a W-EMA strategy.
    After that, the WET results are utilized to distill the MIDN module with rank information through the class-specific ranking distillation (CRD) algorithm.
    Furthermore, WET also acts as a teacher to supervise the  R-CNN head with the multi-seed R-CNN (MSR) algorithm.} 
	\label{framework}
        \vspace{-0.25in}
\end{figure*}

\section{Our Method}
The overall architecture of the proposed framework is shown in Fig.~\ref{framework}. An input image and a set of region proposals are first fed into the basic WSOD module. The proposal features are obtained through a CNN backbone and an RoI Pooling layer followed by two FC layers. Then, these features are fed into the MIDN module to produce instance-level scores. Meanwhile, the image and corresponding proposals are sent to the weighted ensemble teacher (WET) model, which is gradually updated by the basic WSOD module via a W-EMA strategy. After that, the WET results are utilized to distill the MIDN module with rank information through the class-specific ranking distillation (CRD) algorithm. Furthermore, WET acts a teacher to supervise the  R-CNN head with the multi-seed R-CNN (MSR) algorithm.

\subsection{Basic WSOD Module}
\label{sec_wsod}
Due to the lack of instance-level annotations in the WSOD settings, many existing works combine Multiple Instance Learning (MIL) with a CNN model, denoted as Multiple Instance Detection Network (MIDN), to accomplish the detection task. Following previous works \cite{OICR, PCL}, we utilize a two-stream weakly supervised deep detection network (WSDDN) \cite{WSDDN} as our MIDN module.

Given an image $I$, we denote its image-level label as $Y_{img} = [y_1, y_2, \cdots, y_C] \in \mathbb{R}^{C \times 1} $, where $y_c = 1$ or 0 indicates the presence or absence of the class $c$. The generated region proposal set for image $I$ is denoted as $R = \left\{R_1, R_2, \cdots, R_N\right\}$. We first extract proposal features through a CNN backbone and an RoI Pooling layer followed by two FC layers. Then, these proposal features are fed into two sub-branches in MIDN, \textit{i.e.}, classification branch and detection branch. For classification branch, the score matrix $x^{cls} \in\mathbb{R}^{C \times |R|}$ is produced through a FC layer, where $C$ and $|R|$ denote the number of categories and proposals, respectively. Then, a softmax operation is applied on $x^{cls}$ along the categories to produce  $\sigma_{cls}(x^{cls})$.  Similarly, the score matrix $x^{det} \in\mathbb{R}^{C \times |R|}$ is produced in the detection branch by another FC layer. A softmax operation is then applied on $x^{det}$ along proposals to produce $\sigma_{det}(x^{det})$. After that, the classification score for each proposal can be obtained by an element-wise product of these two scores: $x^{midn} = \sigma_{cls}(x^{cls}) \odot \sigma_{det}(x^{det})$. Finally, the image-level classification scores are generated through the the summation over all proposals: $x^{img}_c = \sum_{i=1}^{|R|}x^{midn}_{c, i}$. In this way, we train the MIDN module with binary cross-entropy loss: $\mathcal{L}_{midn}=-\sum_{c=1}^C\left[y_clogx^{img}_c+\left(1-y_c\right)log\left(1-x^{img}_c\right)\right]$

We further follow OICR \cite{OICR} to add several cascaded online instance classifiers (OICs) to refine the classification results. Specifically, for each existing class, OICR selects the top-scoring proposal of the $i$-th classifier and its surrounding ones as positive samples to generate hard pseudo labels $Y_{ref_i} \in\mathbb{R}^{C+1 \times |R|} $. These labels are then used to train the subsequent ($i+1$)-th classifier using a weighted cross-entropy loss $\mathcal{L}_{oic}$. Particularly, pseudo labels of the first classifier OIC$_1$ are generated utilizing the MIDN scores.

Moreover, we add an R-CNN head following \cite{Yang_2019_ICCV, IMCFB}, which consists of two parallel branches for the classification and regression task, respectively. The R-CNN head is supervised by the pseudo labels generated from the last online instance classifier. The weighted cross-entropy loss and smooth-L1 loss are applied for the two tasks, respectively.

\subsection{Weighted Ensemble Teacher}
\label{sec_wet}
In this section, we construct a sibling model of the basic WSOD module, Weighted Ensemble Teacher (WET), to produce more accurate detection predictions. Similar to the former, the WET model consists of a feature extractor and a classification head. An intuitive way to update the WET model is to apply Exponential Moving Average (EMA) following the traditional mean teacher methods \cite{tarvainen2017mean, liu2021unbiased}:
\begin{equation}
\label{EMA}
\theta_{t} \leftarrow \alpha\theta_{t} + (1-\alpha)\theta_{s},
\end{equation}
where  $\theta_{t}$  and $\theta_{s}$ represent the parameters of the same network in the teacher model and the student model, respectively, and $\alpha$ is a smoothing coefficient. Through the EMA strategy, the slowly progressing teacher model can be considered as the ensemble of the student models in different training iterations \cite{liu2021unbiased}. We treat WET and the basic WSOD module as the teacher and student models, respectively. 

However, a problem arises that for the classification head in WET, many candidate networks can be treated as the \textbf{student} models, \textit{i.e.}, $K$ online instance classifiers (OICs) and classification branch (CLS branch) in R-CNN head. The most direct way is to choose CLS (or OIC$_K$) branch as the student network, considering the fact that they always have better performance among these candidates due to their relatively accurate pseudo labels after several refinements. However, this strategy overlooks the potential positive influence of the other candidate student networks.

To this end, we propose to modify EMA to accommodate multiple students. To be specific, we can use the average parameters of all these candidate student networks during EMA (A-EMA) instead of selecting a single candidate:

\begin{equation}
\label{A-EMA}
\theta_{t} \leftarrow \alpha\theta_{t} + \frac{(1-\alpha)}{S}\sum_{s=1}^{S}\theta_{s},
\end{equation}
where $S = K+1$ represents the number of candidates. Nevertheless, assigning the same weight to all the candidates is not the most efficient strategy due to the discrepancies in their performance. To enable the classification head to benefit more from the better candidate, we devise a weighted EMA (W-EMA) to adjust the weight of these candidates accordingly:

\begin{equation}
\label{W-EMA}
\theta_{t} \leftarrow \alpha\theta_{t} + \frac{(1-\alpha)}{2}(\frac{1}{K}\sum_{k=1}^{K}\theta_{k} + \theta_{cls}),
\end{equation}
where $\theta_{k}$ and $\theta_{cls}$ represent the parameters of $k$-th OIC and CLS branch, respectively.  As a result, the weight of CLS branch is amplified to $\frac{K+1}{2}$ of its original value, while the weights of other candidates are decreased. It is noteworthy that W-EMA does not add extra hyperparameters, since it can be viewed as a two-step average of different student parameters (1st for OICs, 2nd for OIC-avg \& CLS).

Overall, we employ EMA for updating feature extractor and W-EMA for updating classification head.
In this paper, we refer to them collectively as W-EMA strategy. The WET model with W-EMA strategy has two main advantages: First, it can reduce the adverse effects of noisy pseudo labels. Second, the WET model can be regarded  as an ensemble model of different student models at different time steps. These advantages enable the WET model to generate more reliable classification results $x^{wet} \in\mathbb{R}^{(C+1) \times |R|}$.

\subsection{Class-Specific Ranking Distillation}
\label{crd}
Given the image-level supervision, MIDN is likely to assign high scores to some erroneous region proposals, such as detecting only the most discriminative parts or containing background noises. To alleviate this problem, we propose to employ additional supervision on the MIDN module. An intuitive way is to generate \textbf{classification-based} supervision with hard pseudo labels, similar to those used for refinement modules \cite{OICR, PCL}. However, this will exceed the limits of the original MIL constraint and, more importantly, contradict our original goal of solving the inaccurate scoring assignment problem of MIDN. For further discussions please refer to Supplementary Material.

To this end, instead of applying supervision for each individual proposal, we turn to distill MIDN with the \textbf{rank-based} information among associated proposals. The most straightforward approach for rank distillation is to drive MIDN to generate a rank distribution similar to that of WET for \textit{all proposals}. However, this approach has two main drawbacks: On one hand, learning the rank distribution among inaccurate samples or irrelevant samples is futile. On the other hand, without instance-level annotations, some positive samples will inevitably be assigned lower scores than some negative ones. To ameliorate  these issues, we design a Class-Specific Ranking Distillation (CRD) algorithm to guide MIDN to adjust to a more appropriate rank distribution among \textit{confident associated proposals}.

Specifically, for an existing object class $c$ (\textit{i.e.}, $y_c = 1$), we first select the proposal with the highest WET score on this class $R_{i_c}$, which is the most confident positive sample. Then, we calculate the overlaps between all proposals with the top-scoring one $R_{i_c}$, and set an overlap threshold $\tau$ to construct a neighboring positive proposal set $P_c$:
\begin{equation}
\label{pos_proposal}
P_c = \{R_i | IoU(R_i, R_{i_c}) > \tau, R_i \in R \}.
\end{equation}
Furthermore, to encourage MIDN to focus on the rank distribution under different views, we continuously increase the overlap threshold $\tau$ with a linear growth strategy:
\begin{equation}
\label{tau}
\tau = \tau_0 + (\tau_1 - \tau_0)\frac{iter_{cur}}{iter_{max}}, \quad \tau\in[\tau_0, \tau_1],
\end{equation}
where $iter_{cur}$ represents the current iteration, and $iter_{max}$ represents the total training iteration using CRD algorithm. $\tau_0$ and $\tau_1$ are set to 0.5 and 1.0 naturally following the common evaluation metrics for selecting positive proposals.

\begin{figure}[t!]
	\centering
	\includegraphics[width=0.47\textwidth, height=0.20\textwidth]{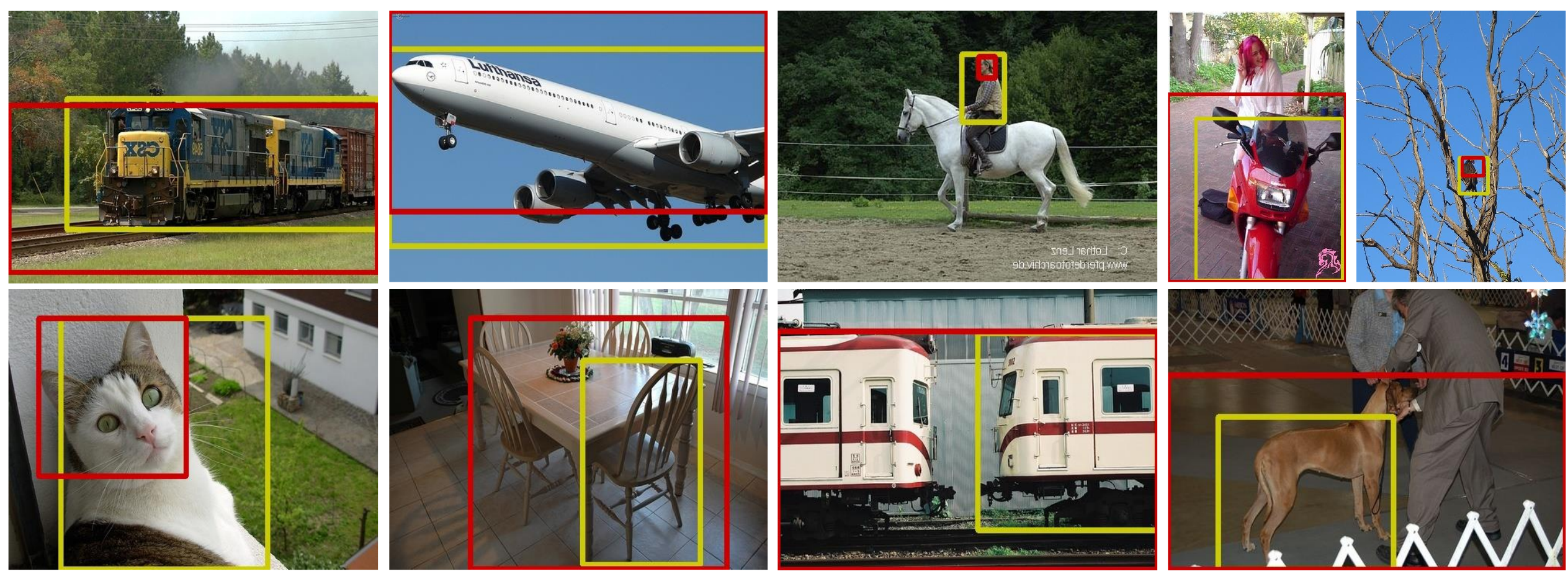} \\
	\caption{Comparison of the top-scoring proposals of the MIDN module in different frameworks at 20$k$ iteration. The proposals from basic WSOD framework are in red and proposals from our CBL framework are in yellow. } 
    \vspace{-0.22in}
	\label{mil}
\end{figure}

After obtaining the positive proposal set $P_c$ for class $c$, we use their predicted scores to represent the rank distribution, since a higher score implies that the corresponding proposal will receive a higher rank in a particular class. We opt for the soft score instead of a hard ranking number since the soft supervision target in distillation is more effective in preserving detailed rank information. Then, a softmax operation is applied on their $c$-th scores for normalization to represent the rank distribution for class $c$:
\begin{equation}
\label{softmax}
s'_{c,j} = \frac{e^{x^{midn}_{cj}}}{\sum_{k=1}^{|P_c|}e^{x^{midn}_{ck}}}, \ t'_{c,j} = \frac{e^{x^{wet}_{cj}}}{\sum_{k=1}^{|P_c|}e^{x^{wet}_{ck}}}, \quad  R_{j}, R_{k} \in P_c,
\end{equation}
where $s'_c$ and $t'_c$ represent the rank distribution of MIDN (student) and WET (teacher) for class $c$.

Finally, we utilize a weighted KL divergence loss to distill the MIDN with rank distributions from WET:
\begin{equation}
\label{kl_loss}
\mathcal{L}_{crd}= -\sum_{c}{\mathbb{I}(y_c =1)\frac{w_c}{|P_c|}\sum_{j=1}^{|P_c|}{t'_{c,j}log(\frac{s'_{c,j}}{t'_{c,j}})}},
\end{equation}
where $w_c$ represents the loss weight of class $c$. 
We apply the highest WET score on this class ($x^{wet}_{ci_c}$) as $w_c$ to represent the confidence of the selected proposal set of this class ($P_c$).

By utilizing  the CRD algorithm, MIDN is encouraged  to adjust higher ranks to more accurate proposals compared with that in the original framework, as shown in Fig.~\ref{mil}.

\subsection{Multi-Seed R-CNN}
\label{msrcnn}
Due to its  good performance, WET model can serve as a reliable teacher to other networks in the basic WSOD module apart from MIDN. In this section, we propose to employ WET for the supervision of the R-CNN head.

The pseudo label generation for R-CNN head in the original WSOD module can be divided into two steps: First, the top-scoring proposal (scores are from the last online instance classifier OIC$_K$) for an existing class is selected as the positive seed for this class. Then, pseudo labels of all proposals are generated according to the overlaps with the positive seed. This procedure guarantees  the quality of the selected seeds, yet overlooks the potential advantages from other possible seeds in the same image.

To this end, we propose a simple Multi-Seed R-CNN (MSR) algorithm to generate more credible seeds by leveraging the reliable WET results.
First, we use the ensemble of the WET results with the results from the original teacher OIC$_K$: $x^{msr} = (x^{wet} + x^{OIC_K}) / 2$.
Next, we propose to narrow the search range of positive seeds. Taking into account the fluctuating distribution of scores during the training stage, we set a soft threshold to accomplish this. To be specific, for each existing class $c$, the threshold $\sigma^{s}_c$ is denoted as the highest $x^{msr}$ of this class multiplied with a factor $\mu^{s}$. To avoid leaving too many proposals, we further set a threshold $\sigma^{n}$ to limit the number of remaining proposals. $\sigma^{n}$ is denoted as the number of whole proposals multiplied with a factor $\mu^{n}$:
\begin{equation}
\label{rcnn_tau}
\sigma^{s}_c = \mu^{s} \max x^{msr}_c, \quad \sigma^{n} = \mu^{n} |R|.
\end{equation}
We first select the top-$\sigma^{n}$ proposals and then filter out the proposals with scores lower than $\sigma^{s}_c$. Then, we apply the Non-Maximum Suppression (NMS) algorithm to the remained proposals and regard the kept ones as positive seeds. 

To further reduce the impact of noisy seeds, we apply the original results (\textit{i.e.}, OIC$_K$ and WET scores) as references to assess their confidence. Specifically, according to each result, we first remove negative proposals using Eq. \ref{rcnn_tau}.  Then, for each seed $i$, we identify if there is one that is very close to it in the remaining proposals. After that, we calculate the proportion  $p_{i,c}$ of such case for seed $i$ according to  all the results. A high proportion indicates that the seed is recognized  as ``positive" by multiple classifiers, thus making it more confident. Finally, the confidence of a seed is obtained as follows:
\begin{equation}
 \label{weight}
w_i = x^{msr}_{i,c} \cdot (1+p_{i,c}^\gamma).
\end{equation}

We generate corresponding pseudo labels according to these positive seeds for the classification branch and regression branch in R-CNN head. We apply weighted cross-entropy loss and weighted smooth-L1 loss to train these two branches, respectively. The confidence $w_i$ is used as the weight and the loss for the R-CNN head $\mathcal{L}_{rcnn}$ is obtained by combining these two losses. For more details, please refer to Supplementary Material.
\subsection{Training Objectives}
\label{tf}
The overall training objective is the combination of MIDN, online instance classifiers (OICs), and R-CNN head, which is elaborated as follows: 
\begin{equation}
\label{total_loss}
\mathcal{L}_{total} = \lambda \mathcal{L}_{midn} + (1-\lambda) \mathcal{L}_{crd} + \sum^{K}_{k=1}{\mathcal{L}_{oic}} + \mathcal{L}_{rcnn},
\end{equation}
where $\lambda$ controls the weight of image-level supervision $\mathcal{L}_{midn}$ and ranking distillation $\mathcal{L}_{crd}$. At the start of training process, MIDN needs to focus on the basic MIL learning for better multi-class classification, while gradually transitioning its focus to adjusting the rank distribution as the training progresses.  To this end, we apply a linear decay strategy to adjust $\lambda$ from 1 to 0 with the increment of training iteration.  Additionally, we start using MSR algorithm at the $0.4 \cdot maxiter$ iteration, considering WET is still in the initial update stage at the beginning of the training procedure.

\begin{table}[t]
	\centering
    \small
	\renewcommand{\arraystretch}{1.1}
	\setlength{\tabcolsep}{1.8mm}{
		\begin{tabular}{l||c|c}
			\specialrule{.15em}{.05em}{.05em}
			Methods & mAP@0.5 & mAP@[.5, .95] \\
			\hline
			PCL \cite{PCL} & 19.4 & 8.5 \\
			MIST \cite{MIST} & 24.3 & 11.4 \\
            CASD \cite{CASD} & 26.4 & 12.8 \\
			\hline
            \textbf{Ours} & \textbf{27.6} & \textbf{13.6} \\
			\specialrule{.15em}{.05em}{.05em}
		\end{tabular}
	}
 \vspace{0.05in}
 \caption{Performance comparison among the state-of-the-art methods with single model on MSCOCO dataset. }
	\vspace{-1em}
	\label{coco}
 \vspace{-0.05in}
 
\end{table}

\section{Experiments and Analysis}
\subsection{Datasets}
We evaluate our method on the prevalent Pascal VOC 2007, Pascal VOC 2012 \cite{voc2007} and MSCOCO \cite{coco} datasets. For VOC 2007 \& 2012 datasets, we train on \textit{trainval} split (5,011 and 11,540 images for VOC 2007 \& 2012), and report the average precision (AP) on \textit{test} set, together with the correct localization rate (CorLoc) on \textit{trainval} set. Only when the Jaccard overlap between the predicted bounding box and the corresponding ground-truth box is above 0.5, the prediction is regarded as a true positive one. For MSCOCO dataset, we train on the \textit{train} split (82,738 images), and test on its \textit{val} split  (4,000 images). During evaluation, we apply two metrics mAP@0.5 and mAP@[.5, .95] following the standard MSCOCO criteria, respectively.

\subsection{Implementation Details}
We follow the common practice \cite{OICR, PCL} to exploit VGG16 \cite{vgg} pre-trained on Imagenet \cite{imagenet} as the backbone network, and to apply Selective Search \cite{SS} for region proposal generation.  We use SGD for optimization, and momentum and weight decay are set to 0.9 and $5 \times {10}^{-4}$ respectively. The learning rate is set to $1 \times {10}^{-3}$ for the first 50K iterations and $1 \times {10}^{-4}$ for the following 20K iterations. We set $\alpha=0.999$, $\gamma=0.4$, $\mu^{s}=0.7$ and $\mu^{n}=0.05$, and $K$ is set to 3 following the common practice.  $iter_{max}$ is set to 80$k$ for VOC 2007 \& 2012 datasets. Following previous works \cite{OICR, MIST}, multi-level scaling and horizontal flipping data augmentation are conducted in both training and testing. Our method is implemented on  PyTorch \cite{pytorch}, and we run all the experiments on an NVIDIA GTX 1080Ti GPU with a batch size of 4.

\begin{table}[t!]
    \small
	\centering
	\setlength{\tabcolsep}{8.1pt}{
	{
			\begin{tabular}{l | c | c}
                \specialrule{.15em}{.05em}{.05em}
				\hline
				Methods & mAP (\%) & CorLoc (\%) \\ \hline
				OICR~\cite{OICR} & 41.2  & 60.6 \\
				PCL~\cite{PCL} & 43.5 & 62.7\\
				C-MIL~\cite{c-mil} & 50.5 & 65.0 \\
				Yang \textit{et al.}  \cite{Yang_2019_ICCV} & 51.5 & 68.0 \\
				C-MIDN~\cite{C-midn} & 52.6 & 68.7 \\
				SLV \cite{SLV} & 53.5 & \underline{71.0} \\
				WSOD$^2$~\cite{WSOD2} & 53.6 & 69.5 \\
				IM-CFB \cite{IMCFB}  & 54.3 & 70.7 \\
				MIST~\cite{MIST} & 54.9 & 68.8   \\
                CASD ~\cite{CASD} & \underline{56.8} & 70.4 \\
				\hline
                \textbf{Ours} & \textbf{57.4} & \textbf{71.8} \\
				\specialrule{.15em}{.05em}{.05em}
		\end{tabular}}
	}
 \vspace{0.05in}
 \caption{Performance comparison among the state-of-the-art methods with single model on PASCAL VOC 2007.}
	\vspace{-1.8em}
	
	\label{voc2007}
	
\end{table}

\subsection{Comparison with State-of-the-arts}
In Tab.~\ref{voc2007}, we present a comprehensive comparison of our proposed method with existing arts with single model on the VOC 2007 dataset. Our method achieves state-of-art performances of 57.4\% mAP and 71.8\% CorLoc, surpassing previous methods by at least 0.6\% and 0.8\%. Our method outperforms recent works \cite{Yang_2019_ICCV, MIST} that directly use original MIDN module to train cascaded refinement modules, since the proposed CRD algorithm improves the scoring assignment on the valuable neighboring positive proposals in MIDN, which benefits the subsequent pseudo labeling, thus raising the upper limit of the whole WSOD performance.  Some methods \cite{C-midn, IMCFB} also aim to deal with the inaccurate scoring assignment issues of MIDN, but their focus is primarily on the part domination problem that high-scoring proposals surround only the discriminative parts. Different from them, our method distills MIDN with rank information from a reliable WET model, which guides MIDN to assign higher scores to accurate proposals among their neighboring ones. Hence, our work can also handle other cases with inaccurate high-scoring proposals (\textit{e.g.}, containing background noises). Furthermore, our method makes the whole framework as a cyclic-bootstrap procedure through the model ensemble (W-EMA strategy) and the rank-information distillation (CRD algorithm). Therefore, our method also performs better than them.

For the MSCOCO dataset, as shown in Tab.~\ref{coco}, our method produces the state-of-art performances of 27.6\% mAP@0.5 and 13.6\% mAP@[.5, .95], outperforming the  best competitor CASD \cite{CASD} by clear margins of 1.2\% and 0.8\%, which also validates the effectiveness of our work. 

\begin{figure*}[t!]
	\centering
    \includegraphics[width=0.95\textwidth,height=0.45\textwidth]{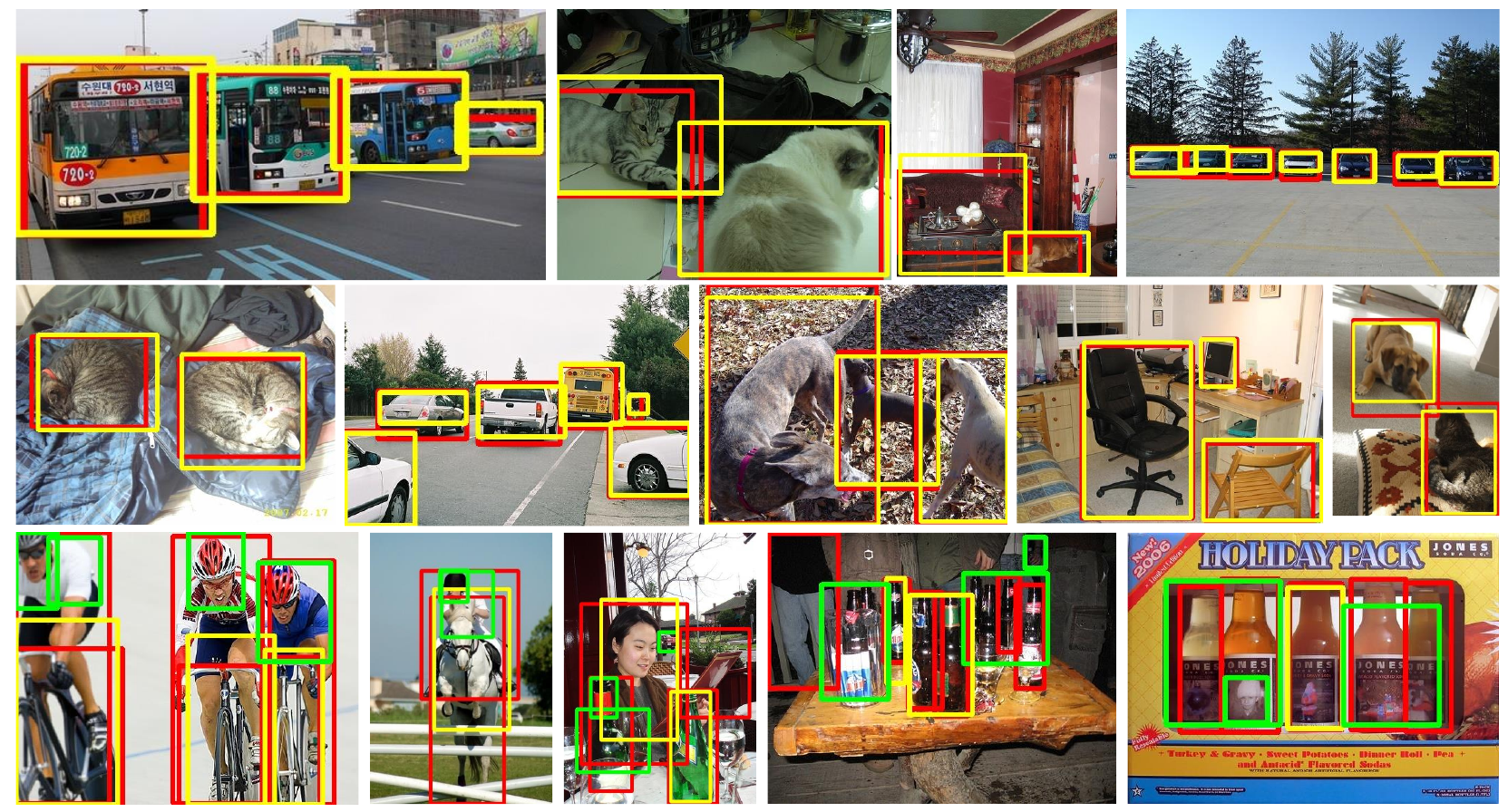} \\%
	\caption{Visualization results on VOC 2007 \textit{test} set. Boxes in red, yellow, and green represent ground-truth boxes, successful predictions, and failure cases, respectively.} 
	\vspace{-0.15in}
	\label{vis_test}
	\setlength{\belowcaptionskip}{-6.5pt} 
\end{figure*}

Fig.~\ref{vis_test} shows the detection results on VOC 2007. The first two rows indicate that our method can detect multiple instances accurately (\textit{e.g.}, ``dog", ``car"), even if they are in some complex scenes. Some failure cases are shown in the last row, which contains localizing only the discriminative parts (\textit{e.g.}, human faces), grouping several objects (especially for ``bottle" class), and containing background parts.

\subsection{Ablation Study}
\subsubsection{Effect of Each Component} We conduct ablation studies on the main components of CBL framework in Tab.~\ref{components} under the mAP metric, where ``inf." represents using WET model during inference. We start from the basic WSOD module (Line 1), with an mAP of 53.3\%. Next, we extend the basic model by adding WET model and use it to distill the MIDN module with CRD algorithm (Line 3), which improves the basic model to 55.8\% mAP, bringing a clear 2.5\% gain.  This outcome highlights the effectiveness of the CRD algorithm in enhancing the overall WSOD performance by distilling rank information on MIDN. After that, we utilize the WET model during inference with the proposed weighted ensemble strategy (Line 4), boosting the performance to 56.0 \% mAP.

To validate the importance of the proposed WET model, we conduct an additional experiment where we replace the WET model with the branch in the basic model  (\textit{i.e.} the last OIC branch OIC$_K$ or the classification branch in R-CNN head) for distillation (Line 2), and we find OIC$_K$ performs better with an mAP of 54.2\%.  However, this operation resulted in a 1.8\% mAP drop, indicating that the WET model is a more dependable teacher in the distillation process. Nonetheless, the performance still remains 0.9\% uperior to the basic model, thus further validating the efficacy of the CRD algorithm. Moreover, when applying the MSR algorithm (Line 5), we can achieve the best performance 57.4\%, which shows that WET can also function as a proficient teacher during the training of the R-CNN head.

\begin{table}[t]
    \centering
    \small
    \setlength{\tabcolsep}{3.00mm}{{
    \begin{tabular}{c | c | c |c | c | c}
    \specialrule{.15em}{.05em}{.05em}
    Basic & WET & CRD & MSR & Inf. & mAP (\%)\\
    \hline
    \checkmark  &  &  &  &  & 53.3 \\
    \checkmark  & & \checkmark &  &  & 54.2 \\
    \checkmark  & \checkmark & \checkmark &  &  & 55.8 \\
    \checkmark  & \checkmark & \checkmark &  & \checkmark & 56.0 \\ 
    \checkmark  & \checkmark & \checkmark & \checkmark & \checkmark & \textbf{57.4} \\ 
    \specialrule{.15em}{.05em}{.05em}
    \end{tabular}
    }}
    \vspace{0.05in}
    \caption{Ablative experiments on the effects of different components in our CBL. The models are evaluated on PASCAL VOC 2007 in terms of mAP (\%).}
    \vspace{-0.1in}
    
    \label{components}
\end{table}

\begin{table}[t]
    \centering
    \small
    \renewcommand{\arraystretch}{1.1}
    	\setlength{\tabcolsep}{3.10mm}{
        {
            \begin{tabular}{l | c}
            \specialrule{.15em}{.05em}{.05em}
            Updating Strategy & mAP (\%)\\
            \hline
            EMA with the last OIC branch & 55.3 \\
            EMA with the CLS branch & 54.5 \\
            A-EMA with all OICs and CLS branch & 56.1 \\
            W-EMA with all OICs and CLS branch & \textbf{57.4} \\
            \specialrule{.15em}{.05em}{.05em}
            \end{tabular}
        }
        }
    \vspace{0.05in}
    \caption{Ablative experiments on the effect of the WET updating strategy. The models are evaluated on PASCAL VOC 2007.}
    \label{abl_wet}
    \vspace{-0.1in}
\end{table}

\begin{table}[t]
    \centering
    \small
    \renewcommand{\arraystretch}{1.1}
    	\setlength{\tabcolsep}{6.2mm}{
        {
            \begin{tabular}{l | c}
            \specialrule{.15em}{.05em}{.05em}
            Overlap Threshold & mAP (\%)\\
            \hline
            Static Value ($\tau=0.75$) & 55.8 \\
            Linear Growth ($\tau \in[0.5, 1.0] $) & \textbf{57.4} \\
            Linear Decline ($\tau \in[0.5, 1.0] $) & 56.2 \\
            \specialrule{.15em}{.05em}{.05em}
            \end{tabular}
        }
        }
    \vspace{0.05in}
    \caption{Ablative experiments on the effect of overlap threshold in CRD. The models are evaluated on PASCAL VOC 2007.}
    \label{abl_tau}
    \vspace{-0.2in}
\end{table}

\subsubsection{Effect of WET Updating Strategy} We conduct experiments to analyze the influence of different updating strategies on WET. The results are shown in Tab.~\ref{abl_wet}, where OIC and CLS represent the online instance classification branch and the classification branch in R-CNN head, respectively. When directly using the single classification branch to update the WET model via EMA (Line 1-2), the proposed WET model can achieve at most 55.3\% mAP, which demonstrates the effectiveness of the whole CBL framework. In addition, utilizing CLS branch does not perform well since it is cascaded after too many refinement modules, hence converging slowly at the beginning of the training procedure. Directly employing  it to update WET will influence the critical initial update phase of WET. When A-EMA is applied (Line 3), the performance shows an improvement of 0.8\% mAP, indicating that other candidates, apart from the best ones, can also have a positive impact on updating the WET model. Additionally, when using the W-EMA strategy (Line 4), the whole framework achieve the best performance 57.4\% mAP. This result shows  that assigning a higher weight to the superior  candidate during the update process is a more effective strategy. 

\begin{figure}[t!]
    \centering
    \includegraphics[width=0.45\textwidth, height=0.17\textwidth]{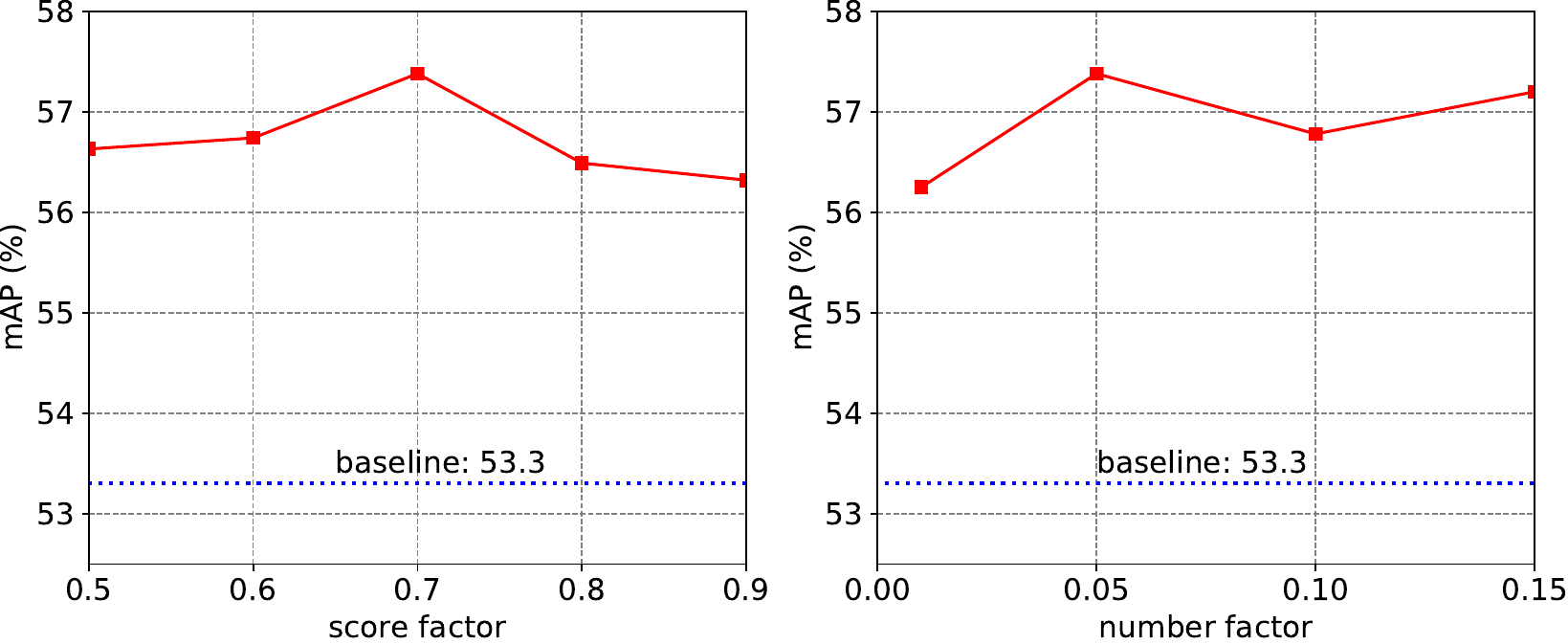} \\
    \caption{Influences of score factor $\mu^{s}$ and number factor $\mu^{n}$ in MSR. The models are evaluated on PASCAL VOC 2007.} 
    \vspace{-3pt}
    \vspace{-1em}
    \label{msr}
    \vspace{-0.05in}
\end{figure}

\vspace{-0.15in}
\subsubsection{Overlap Threshold in CRD}
We compare the different settings on the overlap threshold $\tau$ in CRD, and the results are presented in Tab.~\ref{abl_tau}. We find that changing $\tau$ during training brings more benefits than directly setting a static value, since the former setting will help MIDN to pay attention to the rank distribution under different views. Moreover, a linear growth strategy performs best, because the pseudo labels for the subsequent refinement module need to be more precise with its increasing detection capability. Therefore, CRD algorithm needs to gradually narrow the view to focus on fine-tuning the rank distribution of more accurate proposals.

\vspace{-0.15in}
\subsubsection{Effect of MSR Selection Range} 
\vspace{-0.05in}
Fig.~\ref{msr} shows the influences of score factor $\mu^{s}$ and number factor $\mu^{n}$ used to narrow the selection range of positive seeds. Among all the settings, $\mu^{s} = 0.7, \mu^{n} = 0.05 $ performs best. If the range is too small  (large $\mu^{s}$ or small $\mu^{n}$), few seeds will be found, which limits the benefits from MSR algorithm. Conversely, if the range is too large, some noisy samples will be selected incorrectly, thus degrading the MSR performance. Our MSR algorithm is insensitive to both $\mu^{s}$ and $\mu^{n}$ and all the settings outperform the baseline by at least 3.0\% mAP.

\vspace{-0.1in}
\section{Conclusion}
\vspace{-0.05in}
In this paper, we propose an effective cyclic-bootstrap labeling (CBL) framework for WSOD. We first construct a reliable WET model and update it via W-EMA strategy. After that, the WET results are utilized to distill the MIDN module with rank distribution with the proposed CRD algorithm. Additionally, we propose an MSR algorithm to mine accurate positive seeds to train the  R-CNN head better. The whole framework acts as a cyclic-bootstrap procedure where the subsequent modules of MIDN are finally utilized to supervise itself. Extensive experiments on the PASCAL VOC 2007 \& 2012, and MSCOCO datasets demonstrate the superior performance of our CBL framework.

\textbf{Acknowledgement:} This work was supported by NSFC under Contract U20A20183 and 62021001. It was also supported by the GPU cluster built by MCC Lab of Information Science and Technology Institution, USTC, and the Supercomputing Center of the USTC.

{\small
\bibliographystyle{ieee_fullname}
\bibliography{egbib}
}

\clearpage
\appendix




\section{More Experimental Results}
\subsection{Results on VOC 2012}
In Tab.~\ref{voc2012}, we present a comprehensive comparison of our proposed method with existing arts with single model on the VOC 2012 dataset. Our method achieves a state-of-art CorLoc of 72.6\%, and obtains compatible results on mAP (Ours: 53.5\% vs. SLV: 53.6\%). These results further validate the effectiveness of method.

\subsection{Inference Strategy with WET}
In Tab.~\ref{inf}, we compare different inference strategies with WET scores, where CLS represents the classification branch. We first follow the previous work to only use the basic WSOD module for inference, \textit{i.e.}, averaging the score of $K$ OICs and CLS branch, obtaining an mAP of 57.2\% (Line 1). Then, we add the obtained WET score during the averaging operation, and the mAP is boosted to 57.3\%  (Line 2), justifying the effectiveness of WET. To better utilize the detection capability of WET, we further apply a weighted ensemble strategy and obtain the best performance 57.4\% mAP (Line 3). The strategy can be viewed as a two-step average of different classification results (1st for OICs, 2nd for OIC-avg \& CLS): $x^{inf} = \frac{1}{2}(\frac{1}{K+1} (\sum_{k=1}^{K}{x^{OIC_k}} + x^{cls}) + x^{wet})$, where $x^{cls}$ represents the results of classification branch in the R-CNN head.

Additionally, one may be concerned that  the inference process involving both the basic WSOD module and the whole WET model will cost time. To this end, we apply two strategies to speed up the inference procedure.  One is to directly use WET network for inference, which can also achieve the best performance 57.4\% mAP (Line 4). The other is to discard the feature extractor in WET model during inference (Line 5). In other words, proposal features obtained from the basic WSOD module are directly fed into the CLS branch in the WET model to obtain proposal scores. These WET scores then participate in the averaging operation as mentioned above.  This strategy leads to a 57.3\% mAP, 0.2\% superior to only using the WSOD module.
These results demonstrate that our framework can obtain high performance with negligible extra inference time.

\begin{table}[t!]
    \small
	\centering
	\setlength{\tabcolsep}{7.8pt}{
	{
			\begin{tabular}{l | c | c}
                \specialrule{.15em}{.05em}{.05em}
				\hline
				Methods & mAP (\%) & CorLoc (\%) \\ \hline
				OICR~\cite{OICR} & 37.9 & 52.1 \\
				PCL~\cite{PCL} & 40.6 & 63.2\\
				C-MIL~\cite{c-mil} & 46.7 & 67.4\\
				Yang \textit{et al.}  \cite{Yang_2019_ICCV} & 46.8 & 69.5\\
				WSOD$^2$~\cite{WSOD2} & 47.2 & 71.9\\
				SLV \cite{SLV} & 49.2 & 69.2\\
                    C-MIDN~\cite{C-midn} & 50.2 & 71.2\\
				MIST~\cite{MIST} & 52.1  &  70.9\\
				CASD~\cite{CASD} & \textbf{53.6} & \underline{72.3}\\
				\hline
				\textbf{Ours} & \underline{53.5} & \textbf{72.6} \\
				\specialrule{.15em}{.05em}{.05em}
		\end{tabular}}
	}
    \vspace{0.05in}
    \caption{Performance comparison among the state-of-the-art methods on PASCAL VOC 2012.}
	
    \label{voc2012}
	
\end{table}

\begin{table}[t]
\centering
\small
\renewcommand{\arraystretch}{1.1}
\setlength{\tabcolsep}{2.60mm}{
    {
        \begin{tabular}{l | c}
        \specialrule{.15em}{.05em}{.05em}
        Inference Strategy & mAP (\%)\\
        \hline
        \hline
        Basic WSOD module & 57.2 \\
        \hline
        Basic WSOD + WET score (average) & 57.3 \\
        Basic WSOD + WET score (weighted) & \textbf{57.4} \\
        \hline
        WET score & \textbf{57.4} \\
        Basic WSOD + CLS branch in WET & 57.3 \\
        \specialrule{.15em}{.05em}{.05em}
        \end{tabular}
    }
    }
\vspace{0.05in}
\caption{Ablative experiments on the effects of different inference strategies. The models are evaluated on PASCAL VOC 2007.}
\label{inf}
\end{table}

\begin{table}[t]
\centering
\small
\renewcommand{\arraystretch}{1.1}
	\setlength{\tabcolsep}{2.60mm}{
    {
        \begin{tabular}{l | c}
        \specialrule{.15em}{.05em}{.05em}
        Inference Strategy & mAP (\%)\\
        \hline
        \hline
        Classification head & 56.9 \\
        RoI layer + Classification head & 56.5 \\
        Whole structure & \textbf{57.4} \\
        \specialrule{.15em}{.05em}{.05em}
        \end{tabular}
    }
    }
 \vspace{0.05in}
\caption{Ablative experiments on the effects of different structures of WET. The models are evaluated on PASCAL VOC 2007.}
\label{structure}
\end{table}

\subsection{Effect of different structure of WET}
We conduct experiments using different structures of WET, as shown in Tab.~\ref{structure}. We adopt three strategies to construct WET: Only containing a classification head (Line 1), containing RoI pooling layer and classification head (Line 2), and containing the whole structure (including feature extractor and classification head) (Line 3). We find that the third strategy achieves the best results, since the overall structure can benefit from the EMA strategy to reduce the adverse effects of noisy pseudo labels during training.

\subsection{Effect of confidence rate $\gamma$}
We conduct experiments using various $\gamma$ when generating the confidence of seeds. The results are shown in Tab.~\ref{gamma}. The performance is insensitive to the selection of values near the optimal values we have chosen ($\gamma=0.4$).

\subsection{Effect of EMA rate}
We conduct experiments using various EMA rate $\alpha$. The results are shown in Tab.~\ref{ema}, which indicates that $\alpha=0.999$ is the optimal rate. When the EMA rate is small, the student (Basic WSOD module) contributes more to the teacher (WET model) for each iteration, thus the teacher is likely to suffer from the negative effects brought from the noisy pseudo-labels. When the EMA rate is high, the next model weight of the teacher will be mostly from the previous weight of itself, thus make the teacher grow overly slow. Therefore, we choose $\alpha=0.999$ in our method.

\begin{table}[]
\centering
\small
\renewcommand{\arraystretch}{1.1}
	\setlength{\tabcolsep}{2.5mm}{
    {
        \begin{tabular}{c | c c c c c }
        \specialrule{.15em}{.05em}{.05em}
        $\gamma$ & 0.2 & 0.4 & 0.6 & 0.8 & 1.0  \\
        \hline
        mAP (\%) & 57.2  & \textbf{57.4} & 57.2 & 57.2 & 57.2 \\
        \specialrule{.15em}{.05em}{.05em}
        \end{tabular}
    }
    }
\vspace{0.05in}
\caption{Ablative experiments on the effects of different $\gamma$. The models are evaluated on PASCAL VOC2007.}
\label{gamma}
\end{table}

\begin{table}[]
\centering
\small
\renewcommand{\arraystretch}{1.1}
	\setlength{\tabcolsep}{2.80mm}{
    {
        \begin{tabular}{c | c c c }
        \specialrule{.15em}{.05em}{.05em}
        EMA rate $\alpha$ & 0.99 & 0.999 & 0.9999 \\
        \hline
        mAP (\%) & 55.2 & \textbf{57.4} & 56.4 \\
        \specialrule{.15em}{.05em}{.05em}
        \end{tabular}
    }
    }
\vspace{0.05in}
\caption{Ablative experiments on the effects of different EMA rates. The models are evaluated on PASCAL VOC2007.}
\label{ema}
\end{table}

\subsection{Analysis on MIDN module and OIR$_1$ branch}
Finally, to validate the effectiveness of CRD algorithm, we conduct experiments to evaluate the MIDN module and the OIR$_1$ branch.  Considering the purpose of CRD algorithm to adjust the rank distribution of MIDN module for accurate proposals and the top-scoring strategy with MIDN scores for pseudo labeling, we use mAcc@1 under two strict IoU threholds, (\textit{i.e.}, 0.75 and 0.85), to demonstrate the improvements of MIDN by introducing our proposed CBL. Specifically, for each existing category, we select the top-1 proposal according to the MIDN scores and calculate its overlaps with the ground-truth boxes. The proposal will be regarded as true positive if the maximum overlap is larger than a threshold. Finally, we calculate the Acc@1 for all categories and average them to obtain mAcc@1. 

\begin{figure}[t!]
	\centering
	\includegraphics[width=0.46\textwidth, height=0.15\textwidth]{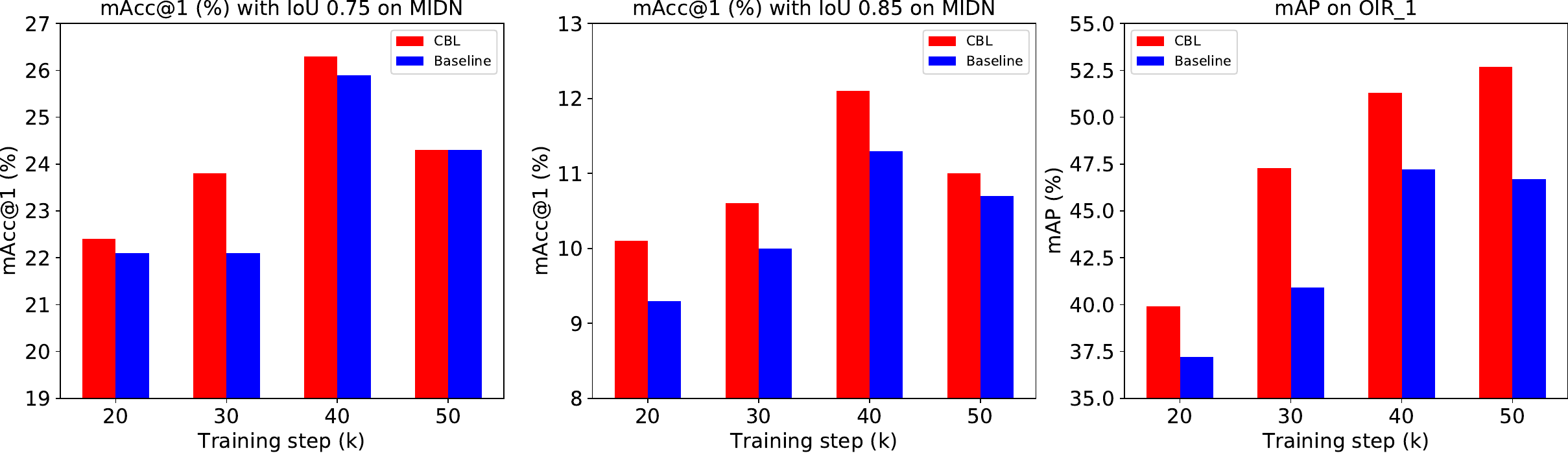} \\
	\caption{Evaluation results for MIDN module and OIR$_1$ branch in different iterations.
	Bars in red and blue represent our CBL framework and the baseline module, respectively.} 
        \vspace{-0.15in}
	\label{res_mil}
\end{figure}

The evaluation results of MIDN module in different iterations are shown in the first two images in Fig.~\ref{res_mil}.  The results show that the MIDN module in our framework outperforms that in the baseline module in most cases. Furthermore, the performance gains are more pronounced during the early stage of training with a loose threshold (0.75), while more evident during the late stage of training with a tight threshold (0.85). This attributes to the linear growth strategy of the overlap threshold in CRD algorithm.  We also conduct experiments on the first OIR branch (OIR$_1$) to show the influence of CRD algorithm on pseudo labeling, since the pseudo labels of OIR$_1$ are generated according to the MIDN scores. The results are shown in the third image in Fig.~\ref{res_mil}. Compared with the baseline module, OIR$_1$ in our framework achieves better mAP performance to a great extend in all cases. 

Overall, with higher mAcc@1 on MIDN module, more seeds close to the ground-truth boxes are successfully chosen in our CBL framework, thus helping generate more high-quality pseudo labels. These accurate pseudo labels will then benefit the training procedure of the OIR$_1$, hence further improving the performance of the whole framework.

\subsection{Additional visualization results}
\begin{figure*}[t!]
	\centering
        \includegraphics[width=0.8\textwidth,height=0.8\textwidth]{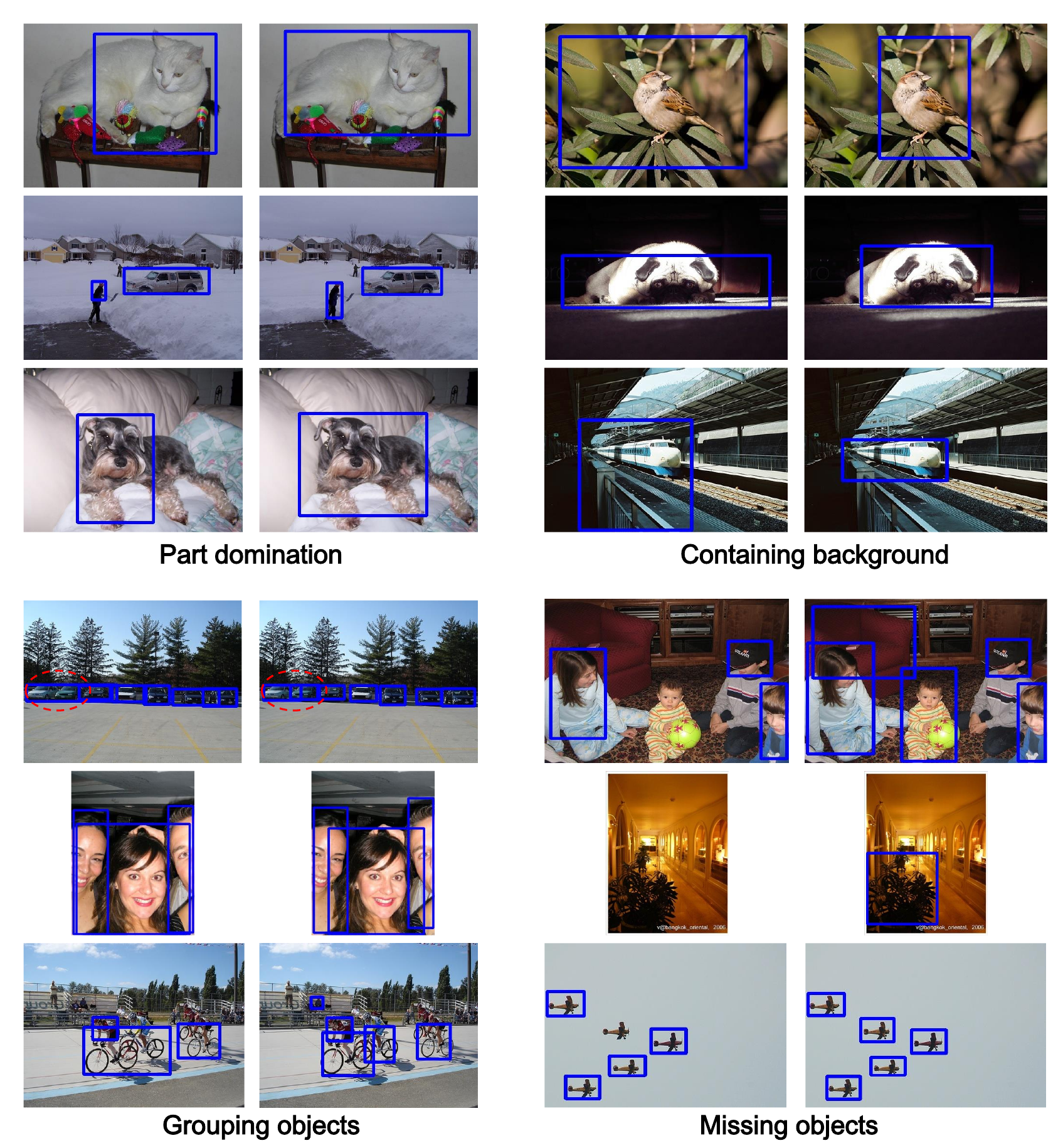} \\
	\caption{Comparison of baseline model and our model. \textbf{Left}: Baseline detection; \textbf{Right}: Our detection. Our method can handle a broader set of inaccurate scoring-assignment cases in baseline detections.} 
	\label{vis_cmp}
\vspace{-0.15in}
	\setlength{\belowcaptionskip}{-6pt} 
\end{figure*}

Fig.~\ref{vis_cmp} compares the detection results of the baseline model  and ours. Benefiting from the cyclic-bootstrap procedure,  our model can handle a broader set of inaccurate scoring-assignment cases, including detecting only discriminative parts (part domination), containing background, grouping objects, and missing objects.

\begin{figure*}[t!]
	\centering
        \includegraphics[width=1.0\textwidth,height=0.65\textwidth]{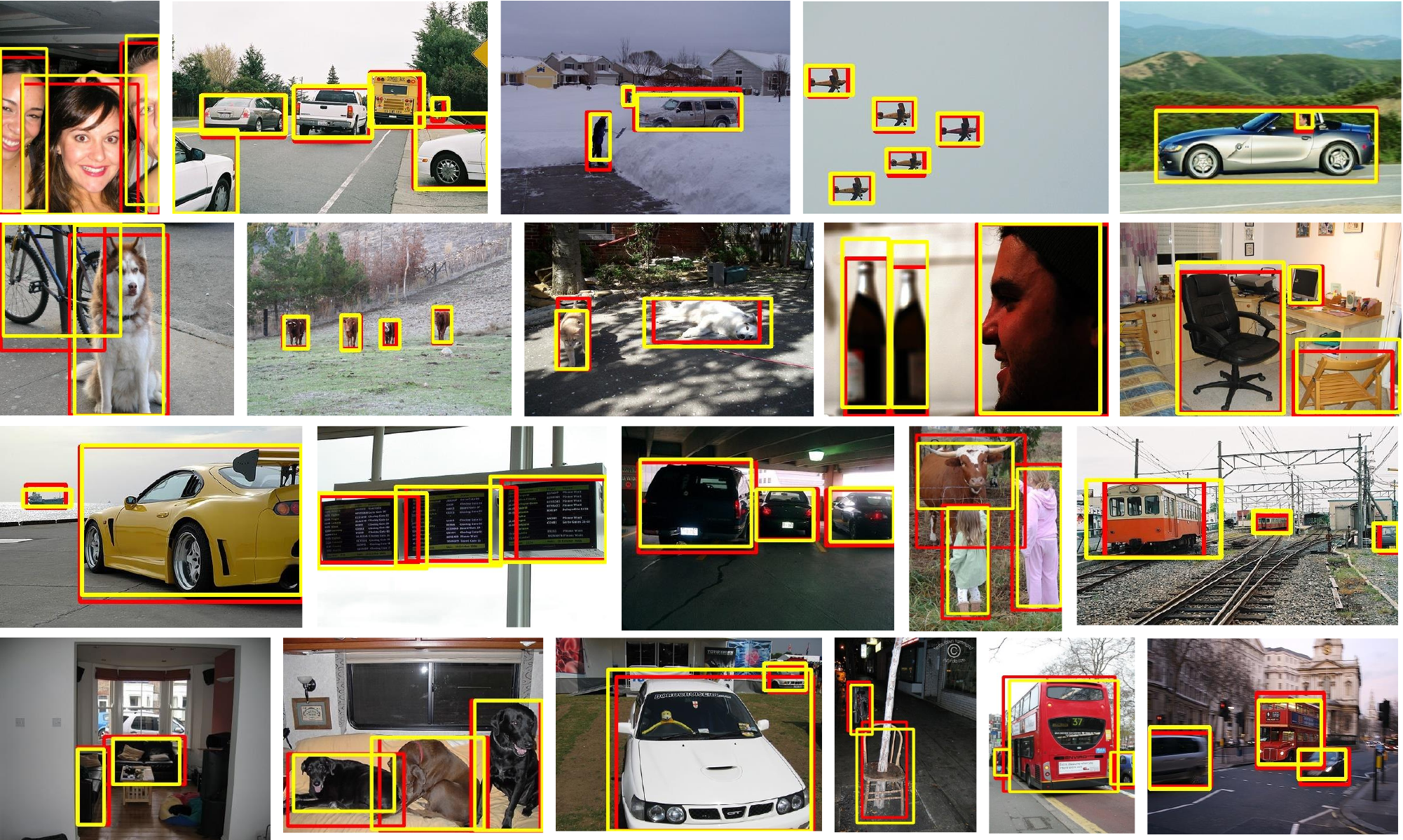} \\
	\caption{Additional visualization resultson VOC2007 dataset. Boxes in red and yellow  represent ground-truth boxes and successful predictions, respectively.} 
	\label{vis_test_more}
	\setlength{\belowcaptionskip}{-6pt} 
 \vspace{-0.1in}
\end{figure*}

Additional visualization results on VOC2007 dataset are shown in Fig.~\ref{vis_test_more}, which demonstrates the detection capability of our method to accurately detect multiple objects (\textit{e.g.}, ``cow", ``plane") in different scenes.

\section{Details of the CBL framework}
\subsection{Softmax operation in MIDN module}
In MIDN module, the softmax operations are different in the classification branch and detection branch, as shown in Eq.\ref{MIL_softmax}:
\begin{equation}
\label{MIL_softmax}
\left\{
\begin{aligned}
\left[\sigma_{cls}\left(x^{cls}\right)\right]_{ij} = \frac{e^{x^{cls}_{ij}}}{\sum_{k=1}^{C}e^{x^{cls}_{kj}}}, \\
\left[\sigma_{det}\left(x^{det}\right)\right]_{ij} = \frac{e^{x^{det}_{ij}}}{\sum_{k=1}^{|R|}e^{x^{det}_{ik}}}.
\end{aligned}
\right.
\end{equation}

\subsection{Loss for the online instance classifiers}
For each online instance classifier, we use weighted cross-entropy loss for training following \cite{OICR}:
\begin{equation}
\begin{split}
\label{loss}
\mathcal{L}_{oic}=-\frac{1}{|R|}\sum_{i=1}^{|R|}\sum_{c=1}^{C+1}w_iy_{c,i}logx_{c,i},
\end{split}
\end{equation}
where $x_{c,i}$ and $y_{c,i}$ represent the predicted OIC score and pseudo label of proposal $i$ on class $c$, respectively. $w_i$ represents the loss weight of proposal $i$, denoted as the corresponding score of its nearest positive seed. $|R|$ and $C$ represent the number of proposals and categories, respectively.

\subsection{Details of the R-CNN head}
For each obtained positive seed, we seek all its neighbor proposals whose overlaps with the seed are greater than 0.5. These neighbor proposals are assigned the same label as their corresponding seed. We regard the selected seeds and their neighbor proposals as positive samples $R_{pos}$, while regarding other proposals as negative ones $R_{neg}$.

For the classification branch, we generate the hard pseudo labels for each proposal $i$: $u_i = [u_{1,i}, u_{2,i}, \cdots , u_{C+1,i}]$. For negative samples, we set $u_{C+1,i} = 1$. Additionally, we ignore the proposals during training whose maximum overlaps with all the seeds are smaller than 0.1. We utilize the weighted cross-entropy loss for training following \cite{OICR}:
\begin{equation}
\label{cls_loss}
\mathcal{L}_{cls}=-\frac{1}{|R|}\sum_{i=1}^{|R|}\sum_{c=1}^{C+1}w_iu_{c,i}logx_{c,i}^{cls},
\end{equation}
where $x^{cls}$ represents the outputs of the classification branch and $w_i$ represents the loss weight of proposal $R_i$ defined in \cite{OICR}. We set $w_i = 0$ for ignored proposal.

For the regression branch, we generate the regression label $v_i = (v_x, v_y, v_w, v_h)$ following \cite{FastR-CNN}. A weighted smooth-L1 loss is utilized for training:
\begin{equation}
\label{reg_loss}
\mathcal{L}_{reg} =-\frac{1}{|R|}\sum_{i=1}^{|R|} \sum_{c=1}^{C} \mathbb I(u_{c, i} = 1)  w_i \cdot \text{smooth}_\text{L1}(t^c_i, v_i),
\end{equation}
where $t \in\mathbb{R}^{(4C) \times |R|} $ represents the outputs of the regression branch. Finally, the loss  for the r-cnn head $\mathcal{L}_{rcnn}$ is obtained by combining these two losses.

\section{Discussion of the supervision on MIDN}
Generating one-hot (hard) labels for each proposal is a more intuitive way to supervise MIDN. However, it has two main disadvantages. On one hand, assigning `1' (foreground) to multiple proposals in the same category will exceed the MIL limitation, where their summation needs to be restricted in [0, 1]. On the other hand, hard labels help correctly classify proposals, but are useless in assigning high classification scores to proposals with more accurate location. Compared with directly assigning hard labels, the CRD algorithm constraints MIDN's prediction to be consistent with the more reliable WET model in the rank distribution of neighboring positive proposals, thus benefiting the scoring assignment of MIDN among them.

\end{document}